\documentclass[runningheads]{llncs}
\usepackage{mytitlesec}
\titlespacing\section{0pt}{12pt plus 4pt minus 2pt}{5pt plus 0pt minus 0pt}
\titlespacing\subsection{0pt}{12pt plus 4pt minus 2pt}{5pt plus 0pt minus 0pt}
\titlespacing\subsubsection{0pt}{12pt plus 4pt minus 2pt}{5pt plus 0pt minus 0pt}
 
\usepackage{eccv}

\usepackage{eccvabbrv}

\usepackage{graphicx}
\usepackage{booktabs}

\usepackage[accsupp]{axessibility}  
\usepackage{multirow}
\usepackage{bm}
\usepackage{xcolor}

\usepackage[pagebackref,breaklinks,colorlinks,citecolor=eccvblue]{hyperref}

\usepackage{orcidlink}

\begin{document}

\title{AMP: Autoregressive Motion Prediction Revisited with Next Token Prediction for Autonomous Driving} 





\author{
Xiaosong Jia$^{1}$\hspace{20pt} Shaoshuai Shi$^{3 \dagger}$\hspace{20pt} Zijun Chen$^{1}$\hspace{20pt} Li Jiang$^{4}$\hspace{20pt} \\
Wenlong Liao$^{1,2}$\hspace{20pt} Tao He$^{2}$\hspace{20pt} Junchi Yan$^{1 \dagger}$
\\
1. Department of CSE, Shanghai Jiao Tong University \\
2. Cowa Robot \hspace{20pt} 3. Max
Planck Institute for Informatics\\
4. The Chinese University of Hong Kong, Shenzhen\\
\normalsize{
$\dagger$ Correspondence authors}\\
}
\institute{ }

\authorrunning{X. Jia et al.}
\titlerunning{Autoregressive Motion Prediction}

\maketitle

\begin{abstract}
As an essential task in autonomous driving (AD), motion prediction aims to predict the future states of surround objects for navigation. One natural solution is to estimate the position of other agents in a step-by-step manner where each predicted time-step is conditioned on both observed time-steps and previously predicted time-steps, i.e., autoregressive prediction. Pioneering works like SocialLSTM and MFP design their decoders based on this intuition. However, almost all state-of-the-art works assume that all predicted time-steps are independent conditioned on observed time-steps, where they use a single linear layer to generate positions of all time-steps simultaneously. They dominate most motion prediction leaderboards due to the simplicity of training MLPs compared to autoregressive networks.

In this paper, we introduce the GPT style next token prediction into motion forecasting. In this way, the input and output could be represented in a unified space and thus the autoregressive prediction becomes more feasible. However, different from language data which is composed of homogeneous units -words, the elements in the driving scene could have complex spatial-temporal and semantic relations. To this end, we propose to adopt three factorized attention modules with different neighbors for information aggregation and different position encoding styles to capture their relations, \emph{e.g.}, encoding the transformation between coordinate systems for spatial relativity while adopting RoPE for temporal relativity.  Empirically, by equipping with the aforementioned tailored designs, the proposed method achieves state-of-the-art performance in the Waymo Open Motion and Waymo Interaction datasets. Notably, AMP outperforms other recent autoregressive motion prediction methods: MotionLM and StateTransformer, which demonstrates the effectiveness of the proposed designs. 
\end{abstract}

\section{Introduction}

Motion forecasting is one key module of a modern autonomous driving (AD) system, which aims to predict the positions of surrounding agents in the next few time-steps so that its downstream module - planner could generate safe route for the ego vehicle without collisions. As the prediction interval could be long (for example, 8 seconds in Waymo Open Motion~\cite{ettinger2021large}), one natural idea is to generate the whole sequence in a step-by-step manner, \emph{i.e.}, autoregressive prediction. Mathematically, autoregressive prediction is defined as: given the observed state $\bm{S}_\text{obs}$ and $t$ already predicted states $\{\bm{S}_\text{1}, ..., \bm{S}_\text{t}\}$, the state at $t+1$ is estimated conditioned both on $\bm{S}_\text{obs}$ and $\{\bm{S}_\text{1}, ..., \bm{S}_\text{t}\}$. Mathematically, for each sampled mode, the whole predicted sequence under \emph{autoregressive setting} with length $T$ is generated as in Eq.~\ref{equ:auto}:
\begin{equation}
\label{equ:auto}
    P(\bm{S}_\text{1}, ..., \bm{S}_\text{T} | \bm{S}_\text{obs}) = \prod\limits_{t=1}^{T} P(\bm{S}_\text{t} | \bm{S}_\text{1}, ..., \bm{S}_\text{t-1}, \bm{S}_\text{obs})
\end{equation}
Since autoregressive prediction fits the intuition that people in the real world move in a \textit{think-act-think-act} circle, pioneering works in the motion prediction field like Social LSTM~\cite{alahi2016social} and MFP~\cite{tang2019multiple} adopt this setting. They combine the recurrent neural network (RNN) and conditional variational auto encoder (CVAE) as their decoder to generate the future trajectory step-by-step. 

However, as pointed out by existing literatures~\cite{jia2023towards}, due to the recurrent nature and the variational inference process of their methods, they suffer from training instability and have difficulty to capture the long-term dependency. As a result, among state-of-the-art works~\cite{shi2022motion,zhou2023query,nayakanti2023wayformer} which dominate widely used large-scale autonomous driving motion prediction datasets~\cite{ettinger2021large,wilson2021argoverse}, they all adopt the simplified assumption that \emph{all future time-steps are independent conditioned on the observed time-steps}. Mathematically, for each sampled mode, the whole predicted sequence under \emph{independent setting} with length $T$ is generated as in Eq.~\ref{equ:indep}:
\begin{equation}
\label{equ:indep}
    P(\bm{S}_\text{1}, ..., \bm{S}_\text{T} | \bm{S}_\text{obs}) = \prod\limits_{t=1}^{T} P(\bm{S}_\text{t} | \bm{S}_\text{obs})
\end{equation}
Compared to autoregressive generation, independent generation could overcome the long-term dependency and optimization issues caused by autoregressive RNN due to the fact that every predicted time-step only depends on $\bm{S}_\text{obs}$ which is input without any cumulative errors. Considering its simplicity and convenience to implement and train, lots of methods are developed based on it including winner solutions~\cite{densetnt,shi2022motion,mercat2020multi,Da2022PathAwareGA,ganet,zhou2023qcnext} of multiple competitions. 

Nevertheless, in the NLP field, the Seq2Seq~\cite{seq2seq1,seq2seq2} model has been eventually replaced by GPT~\cite{radford2019language,gpt3}, partly due to the homogeneous representation of GPT for both inputs and outputs (words) which best describes the inherent characteristics of the language data. In this work, we aim to take a similar step: representing the movement of agents in a unified feature space. Notably, recent  works~\cite{jia2023hdgt,zhou2023query,shi2023mtrpp} have suggested that normalizing the observed states of agents, i.e., inputs, into the ego-centric representation would enhance the performance. Different from previous works, we transform both the observed states and future states into ego-centric tokens and adopt the GPT style next token prediction training. Further, different from pure language data, the autonomous driving scene is heterogeneous, i.e., elements have  multiple types and complex spatial-temporal relations. To this end, we propose multiple tailored designs: (i) We use different tokenizers and attention networks for map and agents. (ii) We use RoPE~\cite{su2021roformer}, the state-of-the-art relative position embedding, to capture the temporal relativity among tokens of same agents. (iii) We encode the transformations between coordinate systems, similar to~\cite{jia2023hdgt,zhou2023query,shi2023mtrpp}, as the relative spatial position encoding for the tokens at the same time-step. After the aforementioned adaptations, the proposed autoregressive motion prediction model, called \textbf{AMP}, achieves state-of-the-art performance on the Waymo Open Motion and Waymo Interaction datasets. In summary, our contributions are:
\begin{itemize}
    \item We propose an autoregressive prediction paradigm for motion forecasting. Under this setting, the inputs and outputs are unified into the same representation space, which could fully utilize the information of whole sequence.
    \item Tailored for the complex spatial-temporal and semantic relations among elements in the driving scene, we design different attention mechanisms and relative position encodings to capture these priors.
    \item The proposed model achieves state-of-the-art performance on the Waymo Open Motion and Waymo Interaction datasets. \textbf{Notably, AMP outperforms other recent autoregressive motion prediction methods: MotionLM and StateTransformer}. 
\end{itemize}

\section{Related Works}
\subsection{GPT Style Modeling}
Transformer~\cite{vaswani2017attention}, with its general design to process information and scalable architecture, has revolutionized the deep learning field. The recent huge succes of ChatGPT~\cite{gpt3,openai2023gpt4} shows the power of next token prediction as a proxy task, which unifies the input and output space. Similar efforts have been made in multiple fields, such as vision~\cite{pmlr-v119-chen20s}, VQA~\cite{openai2023gpt4}, robotics~\cite{zitkovich2023rt}, protein~\cite{ferruz2022protgpt2}, all of which demonstrate strong performance.


\subsection{Motion Prediction}
One intuitive way to conduct motion prediction is to generate the future states in a step-by-step way so that each step is consistent with the previous generated steps. In pioneering learning-based works, they design their decoder based on this intuition. Social LSTM~\cite{alahi2016social} directly fed the predicted step into LSTM~\cite{hochreiter1997long} during training and Social GAN~\cite{gupta2018socialgan} further adds an adversarial recurrent discriminator to enhance diversity. DESIRE~\cite{DBLP:journals/corr/LeeCVCTC17} and MFP~\cite{tang2019multiple} adopt the probabilistic latent variable generative model based on CVAE and their training maximizes a variational lower bound on the log-likelihood of data. However, due to the cumulative error issue and difficulty to optimize~\cite{jia2023towards}, later works~\cite{pmlr-v164-jia22a,mo2021heterogeneous,idenet,LiangECCV20,marchetti2020mantra,ye2021tpcn,hauang2021recoat,gilles2021thomas}  make the assumption that for each mode, all future time-steps are independent conditioned on observed time-steps and thus they could use an MLP to  generate all future states simultaneously. Though counter-intuitive, methods based on this design could outperform the autoregressive prediction model a lot. As a result, state-of-the-art works~\cite{shi2022motion,shi2023mtrpp,zhou2023query} and competition winners~\cite{densetnt,mercat2020multi,Da2022PathAwareGA,ganet,zhou2023qcnext} all follow this design. Notably, SceneTransformer~\cite{ngiam2021scene} treats each input time-step as a token and conducts factorized spatial-temporal attention among observed states of agents. However, they still use an MLP to generate future trajectories. Additionally, since they use a scene-centric representation and thus the representation space of tokens are not unified. Later, there are works to discuss the importance of unified representation, i.e. ego-centric normalization~\cite{jia2023hdgt,zhou2022hivt,shi2023mtrpp}.  Nevertheless, their unification are still only for inputs ignoring the homology between input and output.


\section{Method}
\subsection{Problem Formulation}
The input of the motion prediction task includes (1) the observed states of surrounding agents (vehicles, pedestrians, cyclist, etc) as $\bm{S}_{\text{obs}} \in \mathbb{R}^{N_\text{agent}\times T_{\text{obs}} \times H_{\text{agent}}} $ where $N_\text{agent}$ is the total number of agents, $T_{\text{obs}}$ is the number of observed time-steps, and $H_{\text{agent}}$ is the dimension of recorded state at each time-step (position, velocity, heading, etc) and (2) surrounding map elements $\bm{S}_{\text{map}} \in \mathbb{R}^{N_\text{map}\times L_{\text{map}} \times H_{\text{map}}}$ where $N_\text{map}$ is the number of map elements, $L_{\text{map}}$ is the number of points per polyline, and $H_{\text{map}}$ is the hidden dimension of each point.  

The motion prediction task aims to predict the future states of safety-critical focal agents $\bm{S}_{\text{future}} \in \mathbb{R}^{N_\text{focal}\times T_{\text{future}} \times H_{\text{target}}}$ where $N_\text{focal}$ is the number of selected focal agents, $T_{\text{future}}$ is the number of future time-steps to predict, and $H_{\text{target}}$ is the dimension of the prediction at each future time-step.

\begin{figure}[!tb]
		\centering
  \includegraphics[width=0.99\linewidth]{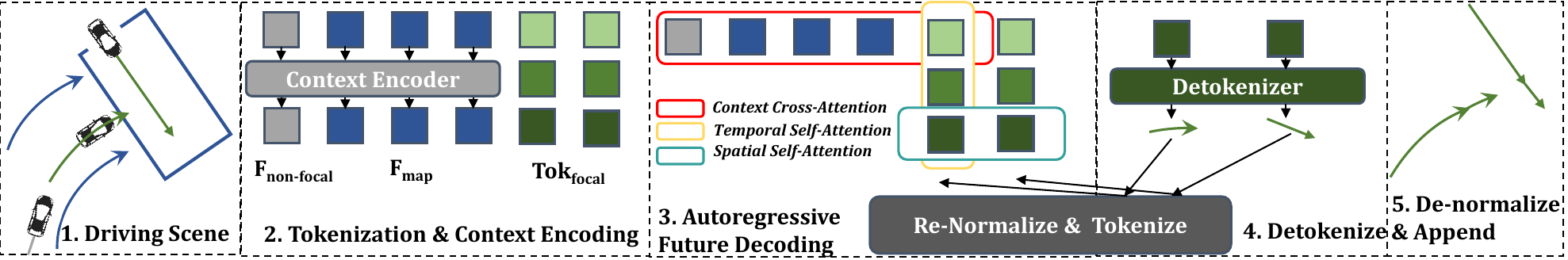}
		\caption{\textbf{Overall Paradigm of AMP.} First, we normalize and tokenize all elements in the scene in Sec.~\ref{sec:tokenization} and use Context Encoder to encode the static map elements and non-focal agents in Sec.~\ref{sec:ce}. Then, we conduct three different attention mechanisms in Future Decoder to autoregressively generate future trajectories in Sec.~\ref{sec:future-decoder}. Next, we detokenize the predicted token into future trajectories in Sec.~\ref{sec:detokenizer}. Finally, we re-normalize and tokenize the predicted trajectories for future generation in Sec.~\ref{sec:inference}. \vspace{-6mm}
  }
\label{fig:overall}
\end{figure}

Different from previous works where they only use the observed states as the inputs of the neural network for training, in this work, we adopt the GPT style next token prediction training strategy. Specifically, during training, both observed and future states of agents are tokenized, then fed into the neural network to perform next token prediction, and finally de-tokenized into predicted states. Fig.~\ref{fig:overall} demonstrates the overall paradigm of the proposed AMP.

\subsection{Discretization \& Tokenization}
\label{sec:tokenization}
\begin{figure}[!tb]
		\centering
  \includegraphics[width=0.85\linewidth]{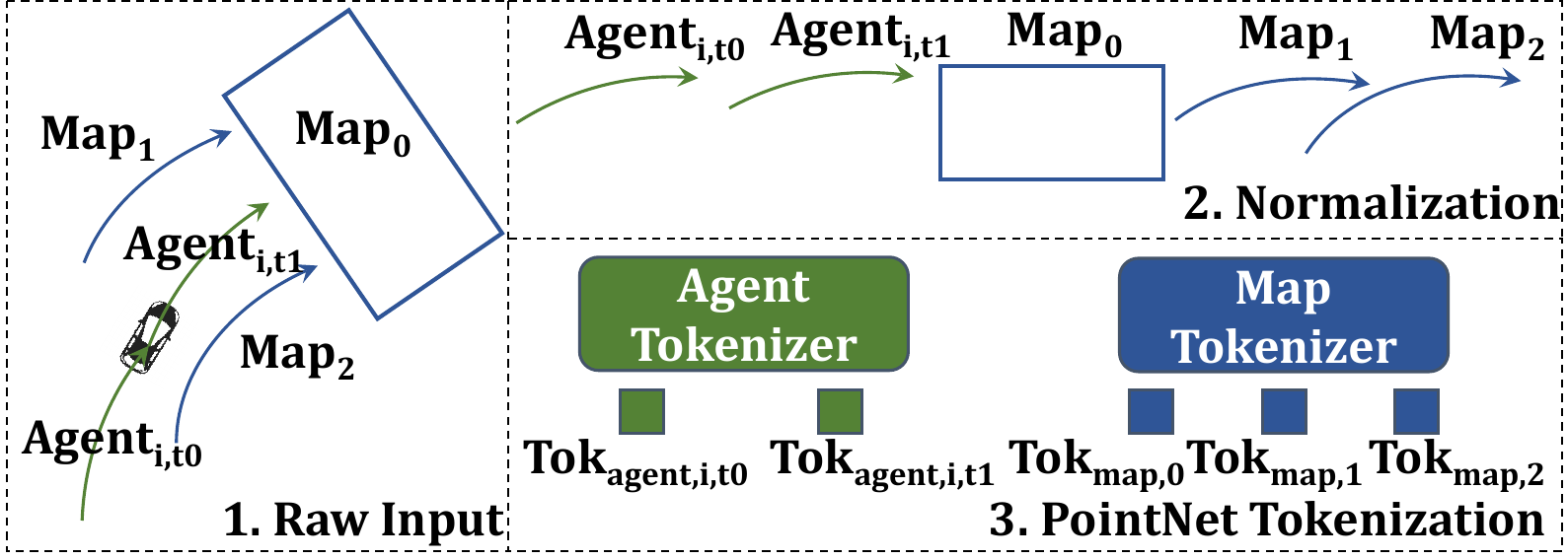}
		\caption{\textbf{Illustration of the Discretization \& Tokenization Process.} All elements are first normalized into ego-centric coordinate system within each small interval and then processed by type-specific PointNet to generate their cooresponding tokens.
  }  \vspace{-5mm}
\label{fig:tokenization}
\end{figure}

In this step, we transform all heterogeneous inputs into a normalized and unified representation space.  We demonstrate the procedure in Fig.~\ref{fig:tokenization}.

\textbf{Map Tokens:} For all map elements, since they are all abstracted as polylines or polygons, we transform their reference coordinate system with origin at the first point and the positive x-axis aligning with the direction from the first to second point. Then, we use a VectorNet~\cite{gao2020vectornet} style PointNet~\cite{qi2017pointnet} to encode the normalized polylines/polygons. We denote the map tokens as $\text{\textbf{Tok}}_\text{map} \in \mathbb{R}^{N_\text{map} \times H }$ were H is the hidden dimension of the model.

\textbf{Agent Tokens:} Different from map elements which are all static, agents usually have a long temporal interval (\emph{e.g.}, in 10 Hz frequency, 1.1 second observed states and 8 second future states in Waymo Open Motion Dataset). Thus, we need to divide each agent into several equally-spaced tokens in the temporal dimension. Considering we only have the observed states as inputs during inference, the temporal length $T_{\text{token}}$ of each token should be shorter than $T_{\text{obs}}$ so that there is at least one token to be the input during inference. As a result, for each agent, we would have $L_{\text{obs}} = T_{\text{obs}}/T_{\text{token}}$ observed tokens and $L_{\text{future}} = T_{\text{future}}/T_{\text{token}}$ future tokens. Next, for each divided small temporal intervals of each agent, we normalize it by setting the reference coordinate system's origin and positive x-axis as the position and heading of agent at the final time-step of the interval. Finally, we use another PointNet shared by all agents and intervals to encode them into observed agent tokens $\text{\textbf{Tok}}_\text{obs} \in \mathbb{R}^{N_\text{agent} \times  L_{\text{obs}} \times H}$  and future agent tokens $\text{\textbf{Tok}}_\text{future} \in \mathbb{R}^{N_\text{focal} \times  L_{\text{future}} \times H}$.  Note that for non-focal agents, i.e., those agents regarded as not interesting or not safety-critical by the datasets~\cite{ettinger2021large,wilson2021argoverse}, we do not calculate their future tokens to save the computation. Since we would only predict focal agents' future, we re-organize these tokens by agents $\text{\textbf{Tok}}_\text{non-focal} \in \mathbb{R}^{N_\text{non-focal} \times  L_{\text{obs}} \times H}$ and $\text{\textbf{Tok}}_\text{focal} \in \mathbb{R}^{N_\text{focal} \times  (L_{\text{obs}}+L_{\text{future}}) \times H}$

In summary, after this step, we would have the tokenized and normalized representations of the scene $\text{\textbf{Tok}}_\text{map}$, $\text{\textbf{Tok}}_\text{focal}$, and $\text{\textbf{Tok}}_\text{non-focal}$.

Different from the homogeneous language data, these tokens have complex spatial-temporal and semantic relations which could not be handled by single self-attention layers. Thus, we adopt a factorized design where tokens attend to different neighbors in different modules instead of being simply put together.

\subsection{Context Encoder}\label{sec:ce}
Since $\text{\textbf{Tok}}_\text{map}$ and $\text{\textbf{Tok}}_\text{non-focal}$ are all fully observable during both training and inference, they could be seen as a \textbf{fixed context}. To reduce the computational burden, \textbf{we pre-compute them with the Context Encoder and then cache them} so that all $\text{\textbf{Tok}}_\text{focal}$ could directly conduct cross-attention to them later. Otherwise, the length of sequence for each focal agent would be much longer (for example, in Waymo Open Motion, there could be thousands of context tokens).

As for the implementation of the Context Encoder, following MTR~\cite{shi2022motion}, we conduct top-k self-attention among $\text{\textbf{Tok}}_\text{map}$ and $\text{\textbf{Tok}}_\text{non-focal}$.  Since all tokens' features are in their own ego-centric coordinate system originally, we need to encode their spatial relations. To this end, we concatenate the relative spatial position encodings~\cite{jia2023hdgt,zhou2023query,shi2023mtrpp} with the token feature maps during attention process. Specifically, assume the references of the coordinate system for elements $i, j$ are $(x_i, y_i, \theta_i)$ and $(x_j, y_j, \theta_j)$ respectively, then the relative spatial relations from $i$ to $j$ is $\Delta_{i\rightarrow j} = (x_j-x_i, y_j-y_i, \theta_j-\theta_i)$. Following NeRF~\cite{mildenhall2020nerf}, we adopts Fourier encoding and then an MLP to encode it as in Eq.~\ref{equ:pos}:
\begin{equation}
\label{equ:pos}
    \text{\textbf{Pos}}_{i;j} = \text{MLP}_{\text{pos}}(\text{Fourier}(\Delta_{i\rightarrow j}))
\end{equation}
where $\text{\textbf{Pos}}_{i;j}$ is the relative spatial position embedding from element $i$ to $j$. 

In summary, the attention in Context Encoder works as in Equ. \ref{equ:ce}:
\begin{equation}
    \label{equ:ce}
    \begin{split}
            \bm{F}_i = \text{MHA}(Q&=[\text{\textbf{Tok}}_i;\text{\textbf{Pos}}_{i;\text{Neightbor(i)}}], \\
            K&=[\text{\textbf{Tok}}_\text{Neightbor(i)};\text{\textbf{Pos}}_{\text{Neightbor(i);i}}], \\
            V&=[\text{\textbf{Tok}}_\text{Neightbor(i)};\text{\textbf{Pos}}_{\text{Neightbor(i);i}}]) \\
    \end{split}
\end{equation}
where $\bm{F}_i$ is the updated feature for token $i$ in $\text{\textbf{Tok}}_\text{map}$ or $\text{\textbf{Tok}}_\text{non-focal}$, MHA denotes the multi-head attention operation in Transformer~\cite{vaswani2017attention}, $[\cdot;\cdot]$ represents the feature concatenation, Neighbor($i$) denotes the top-k spatially nearest neighbors of token $i$, and $\text{\textbf{Pos}}_{i;\text{Neightbor(i)}}$ denotes the relative spatial position embedding matrix between element $i$ and all its neighbors. Note that we also use other structures of Transformer including LayerNorm, residual connection, and feedforward network. After stacking multiple layers of Context Encoder, we cache the updated feature of $\text{\textbf{Tok}}_\text{map}$ and $\text{\textbf{Tok}}_\text{non-focal}$ as $\bm{F}_\text{map}$ and $\bm{F}_\text{non-focal}$.

At the end of the Context Encoder, to provide more supervisions for $\bm{F}_\text{non-focal}$, we also adopts the dense future prediction module in~\cite{shi2022motion}, where we predict a rough future trajectories for all non-focal agents and then concatenate them with $\bm{F}_\text{non-focal}$ followed by an MLP.  

\begin{figure}[!tb]
		\centering
  \includegraphics[width=0.85\linewidth]{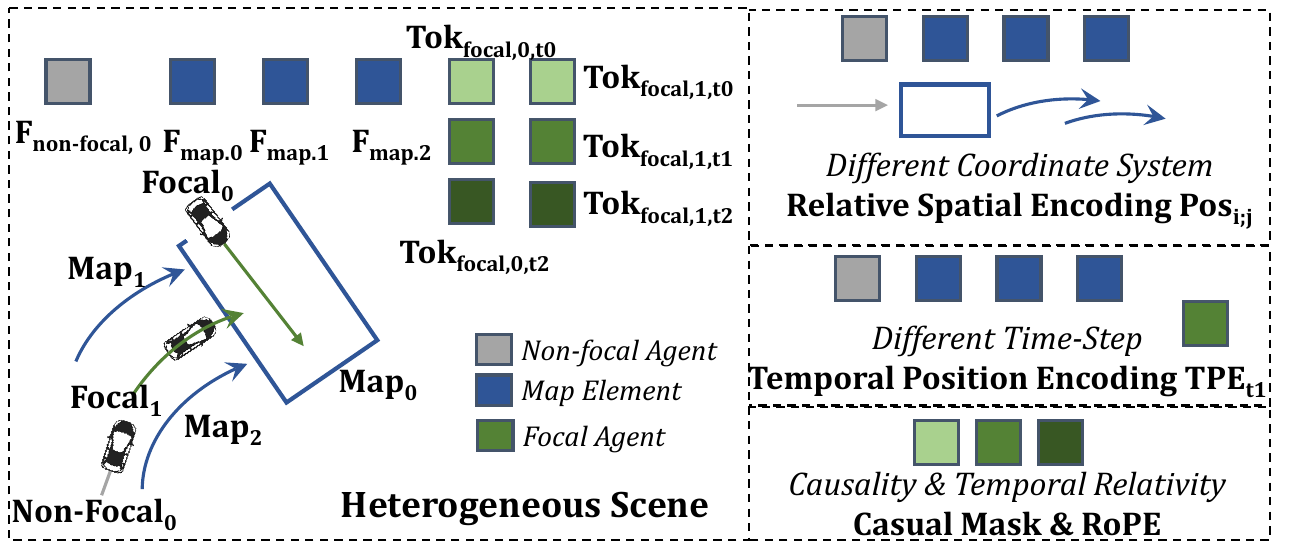}
		\caption{\textbf{Diverse Relations among Driving Scene and Corresponding Position Encodings.} In all attention modules, the coordinate system is different among agents, which requires the Relative Spatial Position Encoding in Eq.~\ref{equ:pos}. For the Context Cross Attention, to let the tokens of focal agents be aware of the temporal difference from the start time-step, we use the Temporal Position Encoding. For the Temporal Self-Attention, to capture the temporal relativity and avoid informantion leakage, we adopt the state-of-the-art designs for GPT - Causal Mask \& RoPE.  \vspace{-6mm}
  }
\label{fig:position}
\end{figure}

\subsection{Future Decoder}
\label{sec:future-decoder}
Future Decoder aims to aggregate the spatial-temporal information among $\text{\textbf{Tok}}_\text{focal}$ and from context tokens $\bm{F}_\text{map}$ and $\bm{F}_\text{non-focal}$. At each layer of the Future Decoder, we employ three attention mechanisms for $\text{\textbf{Tok}}_\text{focal}$ with different parameters, neighbors as introduced in following paragraphs.

\textbf{Context Cross-Attention.} For each focal agent at each time-step, \textbf{their future movement are influenced by the surrounding environments}. In Context Cross-Attention module, we let $\text{\textbf{Tok}}_\text{focal}$ attend to the aforementioned cached and shared context tokens $\bm{F}_\text{map}$ and $\bm{F}_\text{non-focal}$. Note that beside the difference of reference coordinate system mentioned in Sec.~\ref{sec:ce}, there are temporal relativity as well. Specifically, all context tokens are captured at time-step $T_{\text{obs}}$ while $\text{\textbf{Tok}}_\text{focal}$ could be captured at the moment from $T_{\text{obs}}$ to $T_{\text{obs}}+T_{\text{future}}$. To consider this factor, we additionally set a learnable temporal position embedding \textbf{TPE} according to the discretized index of time-steps and add it to the query vector of $\text{\textbf{Tok}}_\text{focal}$.  Thus, the overall attention mechanism of Context Cross-Attention works as in Eq.~\ref{equ:cca}:
\begin{align}
    \label{equ:cca}
    \begin{split}
            \bm{F}_{\text{non-focal},i,t} = \text{MHA}(Q&=[\text{\textbf{Tok}}_{i,t};\text{\textbf{Pos}}_{i,t;\text{non-focal}}] + \text{\textbf{TPE}}_t, \\
            K&=[\text{\textbf{F}}_\text{\text{non-focal}};\text{\textbf{Pos}}_{\text{non-focal;i,t}}], \\
            V&=[\text{\textbf{F}}_\text{\text{non-focal}};\text{\textbf{Pos}}_{\text{non-focal;i,t}}]) \\
    \end{split} \nonumber \\
    \begin{split}
            \bm{F}_{\text{map},i,t} = \text{MHA}(Q&=[\text{\textbf{Tok}}_{i,t};\text{\textbf{Pos}}_{i,t;\text{map}}] + \text{\textbf{TPE}}_t, \\
            K&=[\text{\textbf{F}}_\text{\text{map}};\text{\textbf{Pos}}_{\text{map;i,t}}], \\
            V&=[\text{\textbf{F}}_\text{\text{map}};\text{\textbf{Pos}}_{\text{map;i,t}}] \\
    \end{split} 
\end{align}
where $\bm{F}_{\text{non-focal},i,t}$ and $\bm{F}_{\text{map},i,t}$ are the retrieved features of focal agent $i$ at time-step $t$ about non-focal agents and maps respectively and $\text{\textbf{Pos}}_{i,t;\text{non-focal}}$ and $\text{\textbf{Pos}}_{\text{map;i,t}}$ denote the relative spatial position encoding between focal agent $i$ at time-step $t$ and non-focal agents/maps respectively.

\textbf{Temporal Self-Attention.} During training, each focal agent has $L_{\text{obs}} + L_{\text{future}} = (T_{\text{obs}}+T_{\text{future}})/T_{\text{token}}$ temporal tokens. To enable the GPT style training of \textbf{next token prediction and avoid information leakage}, we apply the causal mask~\cite{radford2019language,gpt3} in the temporal self-attention module. Specifically, during training, each agent's token at time-step $t$ could only attend to the the agent's token before time-step $t$. As for the position encoding, beside the spatial ones used in Eq.~\ref{equ:pos}, there are temporal relativity among tokens as well. We require this position encoding could be the same for the same temporal gap. For example, the relative temporal encoding between time-step $t$ and $t+1$ should be the same with the one between time-step $t+1$ and $t+2$.  To achieve this goal, we adopt the state-of-the-art relative encoding RoPE~\cite{su2021roformer}. Concretely, RoPE is a kind of relative position embedding initially proposed in the NLP field, which combines the sinusoidal encoding with complex number and could be applied on Q and K separately. It is used in both GPT~\cite{gpt3} and LLaMA~\cite{touvron2023llama,touvron2023llama2} and please refer to their paper for details~\cite{su2021roformer}. In summary, the attention mechanism of Temporal Self-Attention is as in Eq.~\ref{equ:tsa}:
\begin{equation}
    \label{equ:tsa}
    \begin{split}
            \bm{F}_{\text{temp}, i, t} = \text{MHA}(Q&=\text{RoPE}([\text{\textbf{Tok}}_{i,t};\text{\textbf{Pos}}_{i,t;i,\{0, ..., t\}}]), \\
            K&=\text{RoPE}([\text{\textbf{Tok}}_{i,\{0, ..., t\}};\text{\textbf{Pos}}_{i,\{0, ..., t\};i,t}]), \\
            V&=[\text{\textbf{Tok}}_{i,\{0, ..., t\}};\text{\textbf{Pos}}_{i,\{0, ..., t\};i,t}], \\
            \text{AttentionMask}&=\text{Causal})
    \end{split}
\end{equation}
where $\text{\textbf{Tok}}_{i,t}$ and $\bm{F}_{\text{temp}, i, t}$ are the input and output features of focal agent $i$ in time-step $t$ for the causally temporal aggregation and $\text{\textbf{Pos}}_{i,t;i,\{0, ..., t\}}$ is the relative spatial position encoding matrix between focal agent $i$'s at time-step $t$ and all time-steps before it.

\textbf{Spatial Self-Attention.} During the selection of focal agents, the dataset builders~\cite{ettinger2021large,wilson2021argoverse} usually combine some rules with manual checking to make sure these agents are interactive or safety-critical. To further \textbf{capture the potential complex interactions among focal agents}, Spatial Self-Attention module considers the spatial relation for focal agents at the same time-step.  However, there is a train-val gap issue due to the multi-modality nature of motion prediction models. 
Specifically, during training, all focal agents are trained by the ground-truth mode, which is unique. During inference, since their future trajectories are generated in a step-by-step manner, each step has multiple possible modes which is different from training and thus it is unknown which mode of agent A should interact with which mode of agent B. To this end, we only conduct Spatial Self-Attention for observable time-steps which have an unique mode during both training and inference. In summary, the Spatial Self-Attention works as in Eq.~\ref{equ:ssa}:
\begin{equation}
    \label{equ:ssa}
    \begin{split}
            \bm{F}_{\text{spatial}, i, t} = \text{MHA}(Q&=[\text{\textbf{Tok}}_{i,t};\text{\textbf{Pos}}_{i,t;\{0, ..., N_{\text{focal}}\},t \}}], \\
            K&=[\text{\textbf{Tok}}_{i,\{0, ..., t\}};\text{\textbf{Pos}}_{\{0, ..., N_{\text{focal}}\},t;i,t}], \\
            V&=[\text{\textbf{Tok}}_{i,\{0, ..., t\}};\text{\textbf{Pos}}_{\{0, ..., N_{\text{focal}}\},t;i,t}])\\
            & \text{where} \quad t \leq t_\text{obs}
    \end{split}
\end{equation}
where $\text{\textbf{Tok}}_{i,t}$ and $\bm{F}_{\text{spatial}, i, t}$ are the input and output of the Spatial Self-Attention for focal agent $i$ at time-step $t$. $\text{\textbf{Pos}}_{i,t;\{0, ..., N_{\text{focal}}\},t \}}$ is the relative spatial position encoding between focal agent $i$ at time-step $t$ and all focal agents at time-step $t$.

\textbf{Layer-Updating MLP.} In the above paragraphs, we introduce three kinds of attention mechanisms in the Future Deocder. In each layer, these attention modules are conducted in parallel. Then, we concatenate all aggregated information - $\bm{F}_{\text{non-focal},i,t}$, $\bm{F}_{\text{map},i,t}$, $\bm{F}_{\text{temp}, i, t}$, $\bm{F}_{\text{spatial}, i, t}$ with the origin feature map $\text{\textbf{Tok}}_{i,t}$ and feed them into an MLP to obtain the input for the next layer as in Eq.~\ref{equ:layer-updating}:
\begin{equation}
\begin{split}
    \label{equ:layer-updating}
    \text{\textbf{Tok}}_{i,t}^{(l+1)} &= \text{MLP}([\text{\textbf{Tok}}_{i,t}^l; \bm{F}^l_{\text{non-focal},i,t}; \\
    & \bm{F}^l_{\text{map},i,t};\bm{F}^l_{\text{temporal}, i, t};\bm{F}^l_{\text{spatial}, i, t}]
\end{split}
\end{equation}
where $l$ is index of layer in the Future Decoder. In Fig.~\ref{fig:position}, we illustrate diverse relations in elements and the corresponding position encodings used for the attention among them.

\subsection{Multi-Modal Detokenizer}
\label{sec:detokenizer}
After Future Decoder, focal agent tokens $\text{\textbf{Tok}}_\text{focal}$ has aggregated spatial-temporal information from environment along with its ego movement. The Multi-Modal Detokenizer generates multiple possible futures based on $\text{\textbf{Tok}}_\text{focal}$. 

Inspired by the intuition that human moves with instant reactions under an ultimate goal, the Multi-Modal Detokenizer composes of \textbf{both short-term and long-term intention prediction and trajecotry generation}. Specifically, given $\text{\textbf{Tok}}_\text{focal, i, t}$, we first distinguish its potential multi-modal long-term goals by concatenating with different embeddings as in Eq.~\ref{equ:long-term}:
\begin{equation}
    \label{equ:long-term}
    \text{\textbf{Tok}}^k_\text{focal, i, t} = [\text{\textbf{Tok}}_\text{focal, i, t}; \bm{F}^k_{\text{long}}] \quad k\in\{1,...,K_{\text{long}}\}
\end{equation}
where $K_{\text{long}}$ is a hyper-parameter for the number of potential long-term goals and $\bm{F}^k_{\text{long}}$ is the learnable embedding for the goal $k$. In this work, we adopt the  state-of-the-art anchor-based design~\cite{shi2022motion} where the long-term goals are generated by k-means clustering of the normalized positions at $T_\text{future}$ over the whole training set. Then, based on its long-term goals, agents should react to current environment. Here, we define the set of potential short-term goals as k-means clustering of the normalized positions at $T_\text{token}$. The short-term goal's label is the one have shortest distance with the ground-truth. Thus, we could jointly train a short-term classifier $\text{CLS}_\text{short}$ and update the $\text{\textbf{Tok}}^k_\text{focal, i, t}$ based on its short-term goals' embeddings $\bm{F}_\text{short}$ as in Eq.~\ref{equ:short}:
\begin{equation}
    \begin{split}
        \text{\textbf{Tok}}^{k\prime}_\text{focal, i, t} &= \text{MLP}_\text{short}([\text{CLS}_\text{short}(\text{\textbf{Tok}}^k_\text{focal, i, t})\cdot\bm{F}_\text{short}; \\ 
        & \text{\textbf{Tok}}^k_\text{focal, i, t}])
    \end{split}
    \label{equ:short}
\end{equation}
where $\text{\textbf{Tok}}^{k\prime}_\text{focal, i, t}$ is the feature containing information about long-term and short-term goals.

\begin{figure}[!tb]
		\centering
  \includegraphics[width=0.99\linewidth]{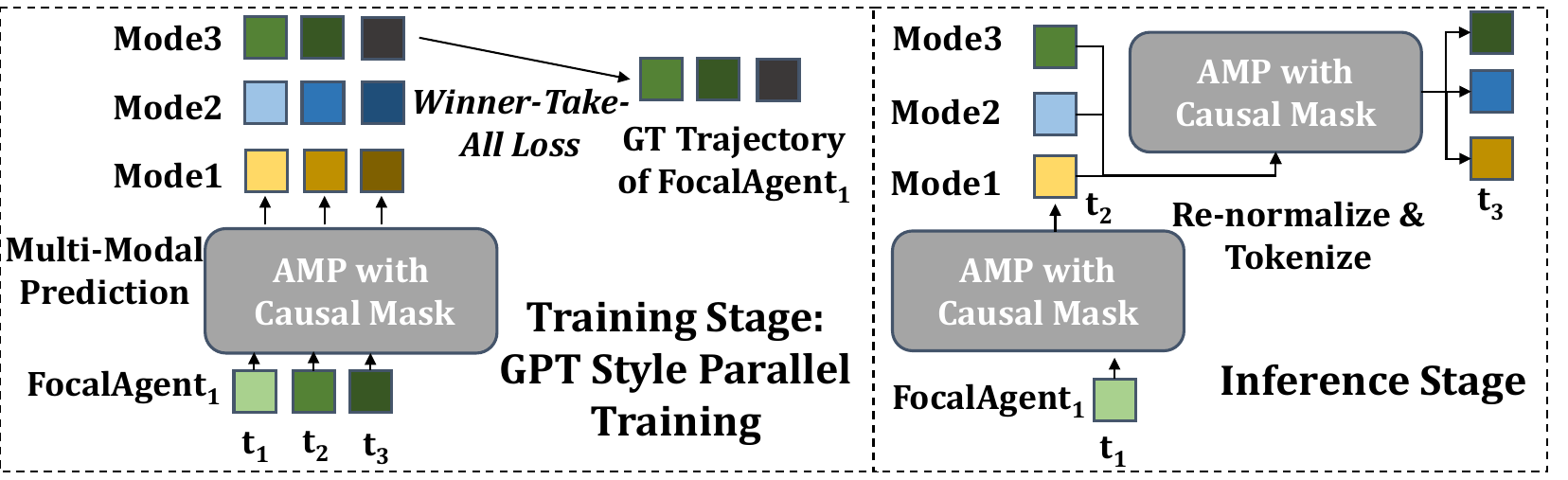}
		\caption{\textbf{Illustration of Training and Inference Stage of AMP.} During training, we implement the GPT-style parallel training for all tokens of the temporal sequence of each focal agent. During inference, the input of the first time-step (observed) is unique while the rest are all autoregressively generated with multiple modes in parallel. \vspace{-6mm}
  }
\label{fig:inference}
\end{figure}

Finally, we use MLPs to decode the $\text{\textbf{Tok}}^{k\prime}_\text{focal, i, t}$ into a concrete trajectory. Here, we still adopt the long-term plus short-term design. We use one MLP $\text{Dec}_\text{short}$ to generate the states in the next $T_\text{token}$ steps $\hat{\bm{O}}^{k}_{i,t,\text{short}}$ and another MLP $\text{Dec}_\text{long}$ to generate the states in the next $T_\text{future}$ steps $\hat{\bm{O}}^{k}_{i,t,\text{long}}$. For inference, we combine the short-term predictions in the current time-step with the long-term predictions in the previous time-steps to generate the predictions $\bm{O}^{k\prime}_{i,t}$ and then re-normalize it into its ego-centric coordinate system so that it could be fed back into the decoder autoregressively. The mechanism of the generation process is:
\begin{equation}
     \label{equ:gen}
    \begin{split}
     \hat{\bm{O}}^{k}_{i,t,\text{short}} &= \text{Dec}_\text{short}(\text{\textbf{Tok}}^{k\prime}_\text{focal, i, t}) \in \mathbb{R}^{T_\text{token} \times H_\text{target}}  \\
     \hat{\bm{O}}^{k}_{i,t,\text{long}} &= \text{Dec}_\text{long}(\text{\textbf{Tok}}^{k\prime}_\text{focal, i, t}) \in \mathbb{R}^{T_\text{future} \times H_\text{target}} \\
     \hat{\bm{O}}^{k}_{i,t} &= \text{Fuse}( \hat{\bm{O}}^{k}_{i,t,\text{short}}, \{\hat{\bm{O}}^{k}_{i,t^\prime,\text{long}}[t-t^\prime] | t^\prime \leq t\}) 
\end{split}
\end{equation}
where Fuse is the fusing function to integrate the current short-term prediction and history long-term prediction, which is weighted averaging in experiments.

Notably, one important advantage of the autoregressive prediction over the independent generation is that it estimates next states based on both existing predictions and the observation $P(\bm{S}_\text{t} | \bm{S}_\text{1}, ..., \bm{S}_\text{t-1}, \bm{S}_\text{obs})$ instead of solely based on the observation $P(\bm{S}_\text{t} | \bm{S}_\text{obs})$. In the experiments, we find that the fusing of short-term and long-term prediction bring performance gains.

\subsection{Objective Function}
Following existing literatures~\cite{shi2022motion}, we adopt the winner-take-all type of loss and decouple the training objectives into regression and classification. Specifically, for both short-term $\hat{\bm{O}}^{k\prime}_{i,t,\text{short}}$ and long-term prediction $\hat{\bm{O}}^{k}_{i,t,\text{long}}$, we only back-propagate the regression loss for one mode, which is selected as the one having shortest distance with the pre-set anchors generated as in Sec.~\ref{sec:detokenizer}. The ground-truth of the classification task is set as the index of the selected mode. The overall loss function is as in Eq.~\ref{equ:loss}:
\begin{equation}
    \label{equ:loss}
    \mathcal{L} = \text{Reg}_\text{short} + \text{Cls}_\text{short} + \text{Reg}_\text{long} + \text{Cls}_\text{long} + \text{Reg}_\text{dense}
\end{equation}
where $\text{Reg}_\text{dense}$ is the dense regression loss mentioned in Sec.~\ref{sec:ce}  and we omit the weight of each term for clarity. Note that all regression loss is calculated based on the normalized labels which are transformed into the ego-centric coordinate system for each agent $i$ at each time-step $t$. 

\vspace{-8pt}
\subsection{Autoregressive Inference}
\label{sec:inference}
During inference time, the initial inputs are only $\bm{S}_{\text{map}}$ and $\bm{S}_{\text{obs}}$ without the future states $\bm{S}_{\text{future}}$. We first use Context Encoder to cache the shared context features $\bm{F}_\text{map}$ and $\bm{F}_\text{non-focal}$. Then, we use Future Decoder and Multi-Modal Detokenizer to generate $K_{\text{long}}$ possible trajectories for next $T_{\text{token}}$ steps. Note that we cache all middle layer features of  Future Decoder $\text{\textbf{Tok}}_{\text{focal}, i,t}^{(l)}$ which could be reused similar to KV-cache~\cite{pope2023efficiently} in GPT. Since the second step, there are $K_{\text{long}}$ temporal sequence decoded for each agent where tokens could only temporally attend to the tokens within the same mode. As for the generated trajectory in each step $\hat{\bm{O}}^{k}_{i,t}$, we need to re-normalize it since the prediction is conducted in the coordinate system at current time-step. Finally, we could feed it into the tokenizer to obtain $\text{\textbf{Tok}}_{\text{focal}, i, t+1}$ to start the generation of next step. The generation halts when reaching the temporal length required by the benchmark. Fig.~\ref{fig:inference} demonstrates the procedure of training and inference stage for comparison.

\section{Experiments}
\subsection{Dataset}
\noindent\textbf{Waymo Open Motion Dataset}:We use the largest open-sourced motion prediction dataset Waymo Open Motion Dataset (WOMD)~\cite{ettinger2021large}. 
For each sample, it gives 11 time-step as observation and the target is to predict the future 80 time-steps.  Within each scene, there are up to 8 focal agents and up to 6 possible futures of each focal agent are allowed for evaluation.  

\noindent\textbf{Waymo Interaction Dataset}: Similar to~\cite{shi2023mtrpp,seff2023motionlm}, we also conduct experiments on the Waymo Interaction Dataset, where all pairs of selected agents are highly interactive and the evaluation is conducted with joint metrics.

\subsection{Metrics}
We use the official metrics suggested by the WOMD and its online leaderboard: minADE, minFDE, MR, and mAP where mAP is the primary metric. MR and mAP measure the ability of covering the focal agent's intentions where minADE and minFDE measure the step-wise errors. All metrics are measured under 2Hz, which means we need to only submit 16 points for the 8 seconds prediction. Please refer to \cite{ettinger2021large} for more details.

\begin{table}[tb!]
\centering
\small
\caption{\textbf{Performance on Waymo Open Motion Dataset.} $\uparrow$ denotes the higher the better while $\downarrow$ denotes the lower the better. Results with ensemble are in \textcolor{gray!50}{\textit{gray}}.} \vspace{-10pt} 
\begin{tabular}{c|l|l|cccc}
\toprule
Split                 & Decoding                  & Methods                 & \textbf{mAP} $\uparrow$   & minADE $\downarrow$ & minFDE $\downarrow$ & MR $\downarrow$     \\ \midrule
\multirow{11}{*}{Test} & \multirow{9}{*}{Indep}    & SceneTransformer~\cite{ngiam2021scene}        & 0.2788 & 0.6117 & 1.2116 & 0.1564 \\ 
                      &                                 & HDGT~\cite{jia2023hdgt}                       & 0.2826 & 0.5703 & 1.1434 & 0.1444 \\
                      &                                 & HPTR~\cite{zhang2023hptr}                       & 0.3904 & 0.5565 & 1.1393 & 0.1434 \\
                      &                                 & \textcolor{gray!50}{\textit{MultiPath++}}~\cite{multipathpp}           & \textcolor{gray!50}{\textit{0.4092}} & \textcolor{gray!50}{\textit{0.5557}} & \textcolor{gray!50}{\textit{1.1577}} & \textcolor{gray!50}{\textit{0.1340}} \\ 
                      &                                 & \textcolor{gray!50}{\textit{WayFormer Factorized}}~\cite{nayakanti2023wayformer}      & \textcolor{gray!50}{0.4118} & \textcolor{gray!50}{0.5447} & \textcolor{gray!50}{1.1255} & \textcolor{gray!50}{0.1229} \\ 
                      &                                 & \textcolor{gray!50}{\textit{WayFormer Multi-Axis}} ~\cite{nayakanti2023wayformer}    & \textcolor{gray!50}{0.4190} & \textcolor{gray!50}{0.5454} & \textcolor{gray!50}{1.1280} & \textcolor{gray!50}{0.1228} \\ 
                      &                                 & MTR~\cite{shi2022motion}                      & 0.4129 & 0.6050 & 1.2207 & 0.1351 \\ 
                      &                                 & MTR++~\cite{shi2023mtrpp}                  & 0.4329 & \textbf{0.5906} & 1.1939 & \textbf{0.1298} \\
                     &                                 &  \textcolor{gray!50}{\textit{MTRA}}~\cite{shi2022mtra}                  & \textcolor{gray!50}{0.4492} & \textcolor{gray!50}{0.5640} & \textcolor{gray!50}{1.1344} & \textcolor{gray!50}{0.1160} \\
                      &                                 & \textcolor{gray!50}{\textit{MTR++\_Ens}}~\cite{shi2023mtrpp}                  & \textcolor{gray!50}{0.4634} & \textcolor{gray!50}{0.5581} & \textcolor{gray!50}{1.1166} & \textcolor{gray!50}{0.1122} \\
                      \cmidrule{2-7}
                      & \multirow{2}{*}{Autoreg} & \textcolor{gray!50}{\textit{MotionLM}}~\cite{seff2023motionlm}               &  \textcolor{gray!50}{0.4364} & \textcolor{gray!50}{0.5509} & \textcolor{gray!50}{1.1199} & \textcolor{gray!50}{0.1058} \\ 
                      &                                 & \textbf{AMP (ours)}          &  \textbf{0.4334}      &    0.6012    &  \textbf{1.1918}      &   0.1311     \\ \midrule \midrule
\multirow{2}{*}{Val}  & \multirow{2}{*}{Autoreg} & StateTransformer~\cite{sun2023large}         & 0.3300   & 0.7200   & 1.4400   & 0.1900   \\ 
                      &                                 & \textbf{AMP (ours)}             & \textbf{0.4321} & \textbf{0.6124} & \textbf{1.2017} & \textbf{0.1352}  \\ \bottomrule
\end{tabular}\vspace{-3mm}
\label{tab:res}
\end{table}

\begin{table}[tb!]
\centering
\caption{\textbf{Performance on Waymo Interaction Dataset.} $\uparrow$ denotes the higher the better while $\downarrow$ denotes the lower the better. Results with ensemble are in \textcolor{gray!50}{\textit{gray}}. For MotionLM in the val set, we choose their best ensemble results and single mdoel with 64 rollouts for fair comparison respectively.} \vspace{-10pt}
\small
\begin{tabular}{c|l|l|cccc}
\toprule
Split                 & Decoding                  & Methods                 & \textbf{mAP} $\uparrow$   & minADE $\downarrow$ & minFDE $\downarrow$ & MR $\downarrow$     \\ \midrule
\multirow{8}{*}{Test} & \multirow{6}{*}{Indep}    & SceneTransformer~\cite{ngiam2021scene}   & 0.1192     & 0.9774 & 2.1892 & 0.4942 \\ 
                      &                                 & M2I~\cite{sun2022m2i}                    & 0.1239   & 1.3506 & 2.8325 & 0.5538  \\
                      &                                 & DenseTNT~\cite{densetnt}                       & 0.1647 & 1.417 & 2.4904 & 0.5350 \\
                      &                                 & \textcolor{gray!50}{JFP}~\cite{luo2023jfp}           & \textcolor{gray!50}{\textit{0.2050}} & \textcolor{gray!50}{\textit{0.8817}} & \textcolor{gray!50}{\textit{1.9905}} & \textcolor{gray!50}{\textit{0.4233}} \\ 
                      &                                 & MTR~\cite{shi2022motion}                      & 0.2037 & 0.9181 & 2.0633 & 0.4411 \\ 
                      &                                 & MTR++~\cite{shi2023mtrpp}                  & \textbf{0.2326} & \textbf{0.8795} & \textbf{1.9505} & \textbf{0.4143} \\
                     &                                 &  GameFormer~\cite{Huang_2023_ICCV}                  & 0.1923 & 0.9721 & 2.2146 & 0.4933 \\
                      \cmidrule{2-7}
                      & \multirow{2}{*}{Autoreg} & \textcolor{gray!50}{\textit{MotionLM}}~\cite{seff2023motionlm}               &  \textcolor{gray!50}{0.2178} & \textcolor{gray!50}{0.8911} & \textcolor{gray!50}{2.0067} & \textcolor{gray!50}{0.4115} \\ 
                      &                                 & \textbf{AMP (ours)}          &  0.2294      &    0.9073    &  2.0415      &   0.4212     \\ \midrule \midrule
\multirow{3}{*}{Val}  & \multirow{3}{*}{Autoreg} & \textcolor{gray!50}{MotionLM}~\cite{sun2023large}         & \textcolor{gray!50}{0.2150}   & \textcolor{gray!50}{0.8831}   & \textcolor{gray!50}{1.9825}   & \textcolor{gray!50}{0.4092}   \\ 
                      &                                 & MotionLM~\cite{sun2023large}             & 0.1585 & 1.0585 & 2.4411 & 0.5114  \\ 
                      &                                 & \textbf{AMP (ours)}             & \textbf{0.2344} & \textbf{0.8910} & \textbf{2.0133} & \textbf{0.4172}  \\ \bottomrule
\end{tabular}\vspace{-6mm}
 \label{tab:interaction}
\end{table}

\subsection{Results}
In Table~\ref{tab:res} and Table~\ref{tab:interaction}, we compare the proposed \textbf{AMP} with state-of-the art works. We could observe that:
\vspace{-2mm}
\begin{itemize}
    \item \textbf{AMP outperforms other recent autoregressive methods.} Both StateTransformer and MotionLM normalize tokens in the agent-level. In AMP, all tokens are in their ego-centric coordinate system and their relativity are encoded by transformations between coordinate systems.  Moreover, MotionLM and StateTransformer both adopt the joint self-attention strategy which flattens all time-steps of all agents into one sequence and thus blurs the temporal relativity. In contrast, AMP adopts a factorized spatial-temporal attention mechanism and the temporal relativity is encoded with state-of-the art relative encoding technique RoPE~\cite{su2021roformer}. As a result, AMP outperforms MotionLM and StateTransformer, which demonstrates the effectiveness of these designs.
    \item \textbf{Gap is narrowed between autoregressive and independent generation.} Compared to the state-of-the-art MTR++~\cite{shi2023mtrpp}, AMP narrows down the gap significantly while MotionLM with ensembles lags behind MTR++\_Ens by 3 mAP in WOMD and 1.5 mAP in Waymo Interaction.
    \item \textbf{Ensemble is still helpful.} By comparing methods with and without ensemble (MTR - MTRA, MTR++ - MTR++\_Ens), we find that ensemble still gives significant performance boosting in the motion prediction field, which demonstrates that current models' performance is far from saturated. 
\end{itemize}


\section{Ablation Studies}
In this section, we study the effectiveness of the proposed designs with small model and 20\% fixed training set of WOMD and we report results on  val set.

\begin{table}[!tb]
\centering
\caption{\textbf{Ablation Study about Position Encodings.}} \vspace{-10pt}
\small
\begin{tabular}{l|cccc}
\toprule
Method                         & \textbf{mAP} $\uparrow$   & minADE $\downarrow$ & minFDE $\downarrow$ & MR $\downarrow$      \\ \midrule
AMP    & \textbf{0.3581}       & \textbf{0.6645} & \textbf{1.4041} & \textbf{0.1717} \\ \midrule
w/o Spatial  & 0.3390       & 0.6968 & 1.4929 & 0.1884 \\
Abs Spatial  & 0.3431       & 0.6899 & 1.4633 & 0.1852 \\
w/o TPE & 0.3520       & 0.6776 & 1.4349 & 0.1762 \\
w/o RoPE                       & 0.3492       & 0.6823 & 1.4475 & 0.1840 \\ \bottomrule
\end{tabular}\vspace{-3mm}
\label{tab:pos}
\end{table}

\textbf{Effects of Different Position Encodings.}
Different from MotionLM and StateTransformer, AMP  has tailored position encodings for different relations among tokens. In Table~\ref{tab:pos}, we explore the effectiveness of them. We could observe that all position encodings are important. (1) The spatial relative position encoding has the largest impact since it influences all attention mechanisms. Without it, each token is unaware of the relative distance from all other tokens and thus leads to large performance drop. Adopting absolute spatial coordinate system still results in performance drop.  (2) Removing the temporal position encoding (TPE) has the least performance degeneration potentially due to the causal mask in the temporal self-attnetion module leaks information about the number of history tokens before current token, which compensate the functionality of TPE. (3) RoPE is useful as well since it brings awareness about the temporal gap between tokens, which could be useful to estimate movement.

\begin{table}[!tb]
\centering
\caption{\textbf{Ablation Study on Training Strategies.} F-BN: freezing the statistics of batch norm layer after epoch 30. N-Focal: using those non-static ones among non-focal agents for training. L-Intention: adopting the local intention feature in Eq.~\ref{equ:short}.}
\vspace{-10pt}
\small
\begin{tabular}{l|cccc}
\toprule
Method                         & \textbf{mAP} $\uparrow$   & minADE $\downarrow$ & minFDE $\downarrow$ & MR $\downarrow$      \\ \midrule
AMP    & \textbf{0.3581}       & \textbf{0.6645} & \textbf{1.4041} & \textbf{0.1717} \\ \midrule
w/o F-BN  & 0.3248       & 0.7028 & 1.5012 & 0.1968 \\
w/o N-Focal & 0.3495       & 0.6898 & 1.4622 & 0.1755 \\
w/o L-Inten    &  0.3492      & 0.6712  &  1.4119 & 0.1801   \\ \bottomrule
\end{tabular}
\label{tab:training}
\end{table}

\textbf{Effects of Training Strategies.}
Since AMP is a new autoregressive prediction paradigm, unlike the independent generation setting which is well-studied in the community, we find some useful strategies which could stabilize the training process and improve the performance. Table~\ref{tab:training} gives results of ablation studies. (i) We could observe that freezing batch norm statistics at the middle of training could bring explicit performance gains. In our early exploration, we find that the loss fluctuates a lot and we conjecture that the GPT structure may not work well with BN since the original ones in the NLP field does not have BN. Thus, we let BN within PointNets and MLPs collect some statistics to enjoy its benefits~\cite{ioffe2015batch} in the early epochs and freeze it later for further convergence. (ii) Adding these non-static (with a movement large than 1 meter) non-focal agents to the training set is beneficial, which is opposite to the observations in the independent generation works~\cite{jia2023hdgt,shi2022motion}. 
(iii) The local intention feature in Eq.~\ref{equ:short} could bring extra supervisions which helps the model to distinguish the multiple potential short-term reactions and thus lead to performance gains.

\begin{table}[!tb]
\centering
\caption{\textbf{Selection of $\tau$ for Fuse Function.}}\vspace{-10pt}
\small
\begin{tabular}{l|cccc}
\toprule
$\tau$                         & \textbf{mAP} $\uparrow$   & minADE $\downarrow$ & minFDE $\downarrow$ & MR $\downarrow$      \\ \midrule
0   &   0.3540     & 0.6722 &  1.4044 & 0.1780 \\ 
0.5  & \textbf{0.3581}       & \textbf{0.6645} & \textbf{1.4041} & \textbf{0.1717} \\
1.0 &  0.3533 & 0.6745  & 1.4102  & 1.1756 \\ \bottomrule
\end{tabular}\vspace{-6mm}
 \label{tab:fuse}
\end{table}

\textbf{Effects of Fuse Function.}
As in Eq.~\ref{equ:gen}, the proposed method generates both the short-term $\hat{\bm{O}}^{k}_{i,t,\text{short}}$ and long-term predictions  $\hat{\bm{O}}^{k}_{i,t^\prime,\text{long}}$. Thus, during inference, the final trajectories could be obtained by fusing the long-term estimation from previous time-steps and the short-term predictions at current time-step. We adopt weighted averaging as in Equ~\ref{equ:avg}:
\begin{equation}
    \label{equ:avg}
    \hat{\bm{O}}^{k}_{i,t} =  t^\tau \hat{\bm{O}}^{k}_{i,t,\text{short}} +  \sum\limits_{t^\prime=0}^t \frac{t^{\prime \tau}}{t^\tau+\sum\limits_{t^\prime=0}^t t^{\prime\tau}} \hat{\bm{O}}^{k}_{i,t^\prime,\text{long}}[t-t^\prime] 
\end{equation}
where $t$ and $t^\prime$ are the temporal index of current time-step and its previous time-steps respectively, $\tau$ is the hyper-parameters where larger $\tau$ indicates more weights for short-term predictions. Table~\ref{tab:fuse} shows that the optimal value for $\tau$ is the 0.5, which demonstrates the importance of both sides.  In future work, more complex fusing functions such as Kalman filter may bring further gains.


\section{Implementation Details}
We implement the model with Pytorch~\cite{pytorch} and the major model has 6 Context Encoder layers and 6 Future Encoder layers with a hidden dimension 512 for fair comparisons with state-of-the-art works~\cite{jia2023hdgt,shi2022motion,shi2023mtrpp} while we use a smaller model with 3 Future Encoder layers with a hidden dimension 256 for all ablation studies to save computational resource. Following~\cite{shi2022motion,shi2023mtrpp}, we generate 64 modes and use the same NMS code to obtain the final 6 trajectories. The model is trained 
for 50 epochs with an initial learning rate 1e-4, weight decay 5e-3, and clip gradients 1.0. We decay the learning rate by half at epoch 35, 40, 45, 49, 50.  Following common GPT training practice, we use Dropout~\cite{JMLR:v15:srivastava14a} and Droppath~\cite{huang2016deep} rate 0.1 and disable them at the last 2 epochs for convergence. Additionally, we find that freezing the batch norm layer in the middle epochs (epoch 30 in our experiments) could be helpful for training stability, possibly due to our small batch size (128 in our experiments) limited by the computational resource. As for focal agents, we find that training with those non-static agents could bring  performance gains, which is quite different from the observation in the previous works~\cite{jia2023hdgt,shi2022motion}. Thus, during training, we randomly select up to 16 non-static non-focal agents into the training set. For submission to the Leaderboard, we combine train, val, and train20s while for ablation studies we use 20\% train set for fast validation.

\section{Visualization}
\begin{figure*}[!tb]
		\centering
		\begin{subfigure}[t]{0.5\textwidth}
			\centering
		\captionsetup{width=7.0cm}
		\includegraphics[width=7.0cm]{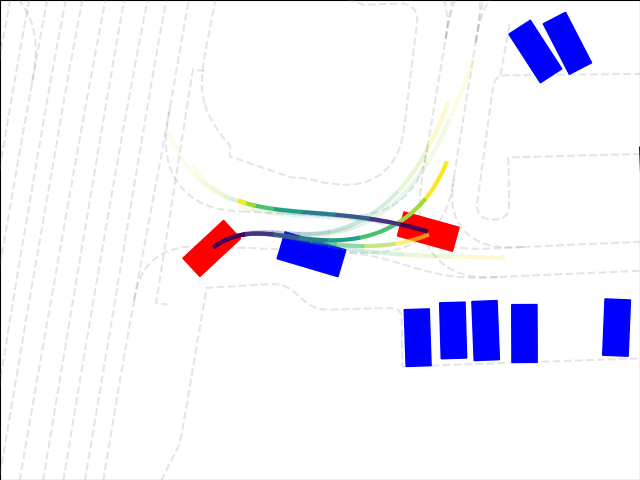}
		\captionsetup{width=7.0cm}
			\caption{Scene ID: 13cdfcfce53ac473. \label{fig:vis1}}
		\end{subfigure}%
		\begin{subfigure}[t]{0.5\textwidth}
			\centering
	   	\includegraphics[width=7.0cm]{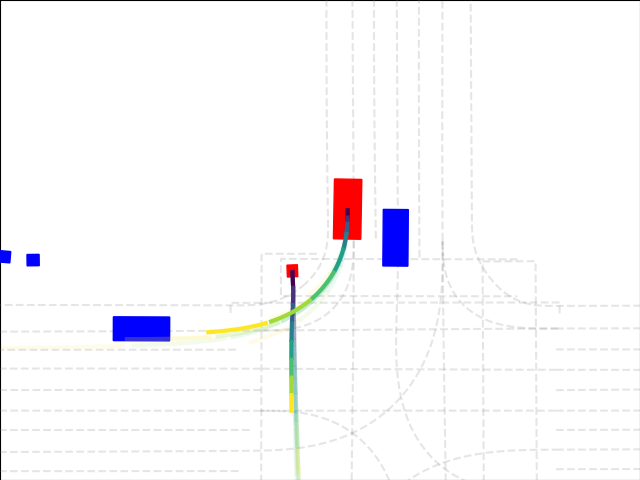}
		\captionsetup{width=7.0cm}
			\caption{Scene ID: 380cf482c5a0ecd9.\label{fig:vis2}}
		\end{subfigure}%
		\\
		\begin{subfigure}[t]{0.5\textwidth}
			\centering
			\includegraphics[width=7.0cm]{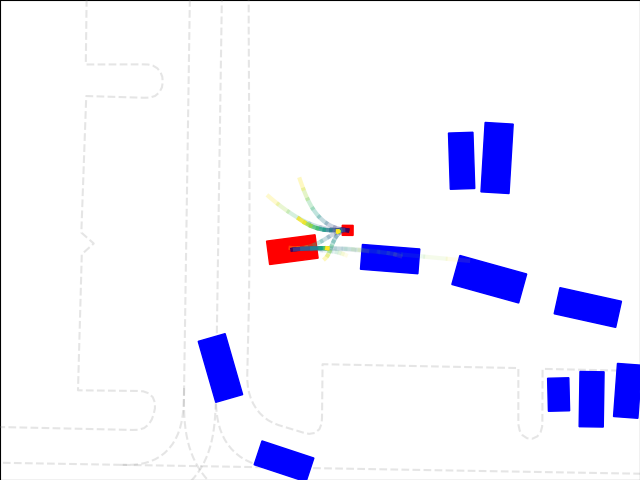}
			\captionsetup{width=7.0cm}
			\caption{Scene ID: 47f46748fe48bc30.\label{fig:vis3}}
		\end{subfigure}%
		\begin{subfigure}[t]{0.5\textwidth}
			\centering
		\captionsetup{width=7.0cm}
		\includegraphics[width=7.0cm]{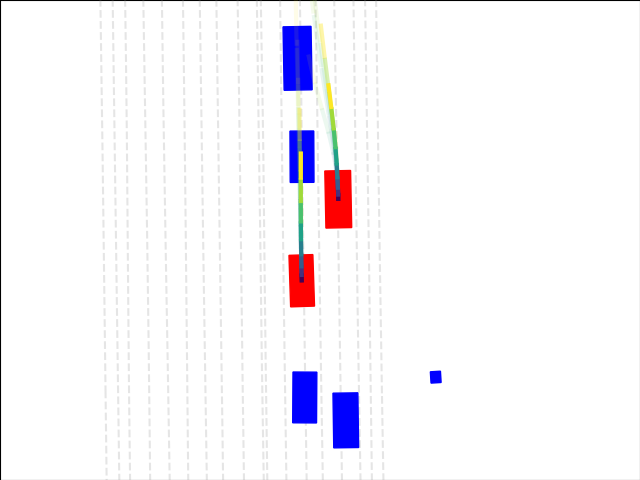}
			\caption{Scene ID: 8a8119876fa94955.\label{fig:vis4}}
		\end{subfigure}%
		\caption{\textbf{Visualization of Prediction Results on Validation Set of Waymo Interaction Dataset.} We use tokens in different timesteps are in different colors. Modes with higher confidence score would be brighter.}
\label{fig:vis}
\end{figure*}

We visualize the multi-mode autoregressive prediction results on the Waymo Interaction Dataset in Fig.~\ref{fig:vis}.  We could observe that autoregressive prediction generates intention consistent results. Specifically, in Fig.~\ref{fig:vis1}, regarding the left vehicle's different target, the right vehicle has different speeds. In Fig.~\ref{fig:vis2}, the vehicle's speed depends on the pedestrian's speed while in Fig.~\ref{fig:vis3} the vehicle's speed depends on the pedestrian's goal point. In Fig.~\ref{fig:vis4}, the two vehicles involved in the cut-in scenario have strong interactions and their future are highly dependent on others.

\section{Conclusion}
In this work, we propose AMP, an autoregressive motion prediction paradigm based on GPT style next token prediction training, where observed and future states of agents are unanimously represented as tokens in the ego-centric coordinate system. To deal with the complex spatial-temporal and semantic relations among tokens, factorized attention mechanisms with tailored position encodings are proposed. By combining these designs, AMP achieves SOTA performance

\noindent\textbf{Limitations}: Currently, autoregressive prediction's performance still has small gap with independent generation. Further decoding techniques incorporating classical state estimation methods like Kalman filter could be explored.



\clearpage  

%
%
\bibliographystyle{splncs04}
\bibliography{main}

\end{document}